\documentclass[conference]{IEEEtran}
\IEEEoverridecommandlockouts
\renewcommand{\thesection}{\Roman{section}}

\usepackage{amsmath, amssymb, amsfonts}
\usepackage{graphicx}
\usepackage{textcomp}
\usepackage{cite}
\usepackage{booktabs}
\usepackage{multirow}
\usepackage{tabularx}
\usepackage{array}

\usepackage{algorithm}
\usepackage{algorithmic}

\usepackage[usenames, dvipsnames]{xcolor}
\definecolor{myblue}{RGB}{10, 150, 200}
\definecolor{highlightColor}{HTML}{E6FFE6}

\usepackage{caption}
\usepackage{adjustbox}
\usepackage{fancyhdr}
\usepackage{comment}

\usepackage[hidelinks]{hyperref}

\def\BibTeX{{\rm B\kern-.05em{\sc i\kern-.025em b}\kern-.08em
    T\kern-.1667em\lower.7ex\hbox{E}\kern-.125emX}}

\title{Improving Fine-Grained Rice Leaf Disease Detection via Angular-Compactness Dual Loss Learning}

\author{
    \IEEEauthorblockN{
        Md. Rokon Mia\textsuperscript{1},
        Rakib Hossain Sajib\textsuperscript{2},
        Abdullah Al Noman\textsuperscript{3},
        Abir Ahmed\textsuperscript{4},
        B M Taslimul Haque\textsuperscript{5}
    }
    \IEEEauthorblockA{
        \textsuperscript{1,2}Department of Computer Science and Engineering, Begum Rokeya University, Rangpur, Rangpur, Bangladesh\\
        \textsuperscript{3}Wilmington University, New Castle, Delaware, United States\\
        \textsuperscript{4}Washington University of Science and Technology, Alexandria, Virginia, United States\\
        \textsuperscript{5}Central Michigan University, Mount Pleasant, Michigan, United States\\ 
        Email: miarokon2001@gmail.com,
        rakibnsajib@gmail.com,
        anoman001@my.wilmu.edu,\\
        abira.student@wust.edu,
        haque2b@cmich.edu
    }
}

\begin{document}



\maketitle
\thispagestyle{fancy} 
\fancyhf{} 

\renewcommand{\headrulewidth}{0pt}
\renewcommand{\footrulewidth}{0pt}

\begin{abstract}
Early detection of rice leaf diseases is critical, as rice is a staple crop supporting a substantial share of the world’s population. Timely identification of these diseases enables more effective intervention and significantly reduces the risk of large-scale crop losses. However, traditional deep learning models primarily rely on cross entropy loss, which often struggles with high intra-class variance and inter-class similarity, common challenges in plant pathology datasets. To tackle this, we propose a dual-loss framework that combines Center Loss and ArcFace Loss to enhance fine-grained classification of rice leaf diseases.  The method is applied into three state-of-the-art backbone architectures: InceptionNetV3, DenseNet201, and EfficientNetB0 trained on the public Rice Leaf Dataset. Our approach achieves significant performance gains, with accuracies of 99.6\%, 99.2\% and 99.2\% respectively. The results demonstrate that angular margin-based and center-based constraints substantially boost the discriminative strength of feature embeddings. In particular, the framework does not require major architectural modifications, making it efficient and practical for real-world deployment in farming environments.

\end{abstract}


\section{Introduction}
Rice is a essential global food staple, however its yield is severely threatened by several leaf diseases (e.g., Bacterial Leaf Blight, Brown Spot, Leaf Blast, Leaf Scald, and Narrow Brown Spot), which may result in crop losses ranging from $10\%$ to over $60\%$ \cite{Kumar2023Rice, Baite2020Disease}. Recent advances in deep learning have revolutionized agricultural automation, enabling highly effective image-based diagnosis of plant diseases. Convolutional Neural Networks (CNNs), including architectures such as ResNet, DenseNet, EfficientNet, VGG, and InceptionNet, have shown strong performance in these classification tasks\cite{Nareshkumar2025, Mukherjee2025, Morol2022,Goh2025}.

\begin{figure}[htbp]
    \centering
    \includegraphics[width=\linewidth]{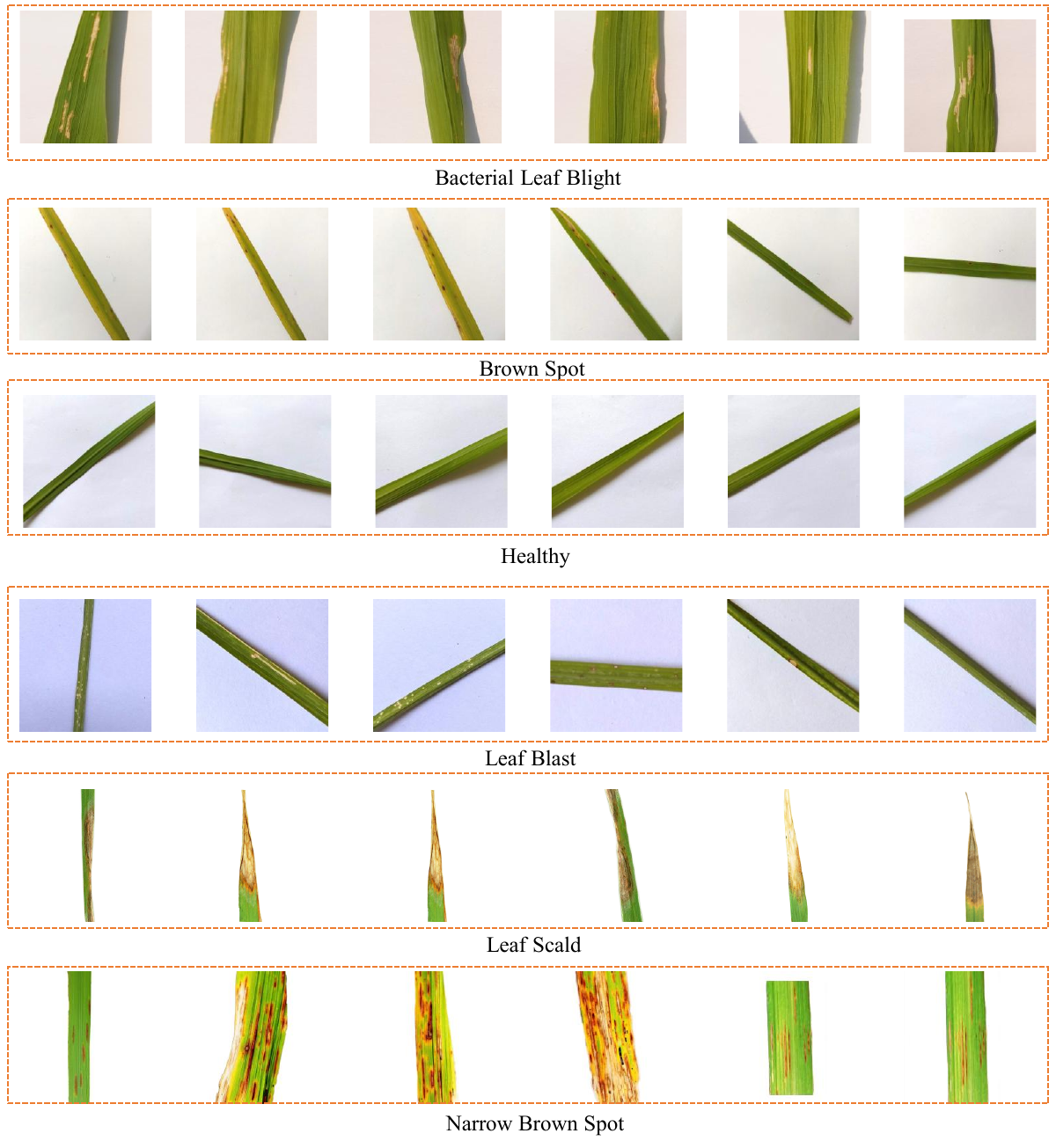}
    \caption{Representative samples of rice leaf disease classes.}
    \label{fig:sample_images}
\end{figure}

However, a significant challenge remains: fine-grained classification of rice leaf diseases is difficult due to subtle and visually overlapping symptoms—for instance, Bacterial Leaf Blight and Leaf Blast, or Brown Spot and Narrow Brown Spot, can exhibit highly similar visual patterns (see Fig.\ref{fig:sample_images}). Most current deep learning methods rely primarily on the cross-entropy loss function. This function is limited because it fails to explicitly enhance intra-class compactness or inter-class separability, which are crucial properties for accurately discriminating between visually similar disease classes.

To overcome this limitation, metric learning-based loss functions, such as Center Loss \cite{Wen2016} (which minimizes intra-class variance) and ArcFace\cite{Deng2018ArcFace} (which maximizes inter-class separation with an angular margin penalty), have proven highly effective in domains like face recognition \cite{Modak2022}. Triplet loss \cite{Hoffer2015TripleteLoss}, particularly is frequently employed in person-reidentification and gait recognition, uses techniques such as increasing intra-class gap and decreasing inter-class gap. Despite their success, these methods remain underexplored in agricultural vision.This study introduces a dual-loss framework combining both ArcFace and Center Loss atop ImageNet-pretrained CNNs (e.g., InceptionNetV3, DenseNet201, EfficientNetB0). Our approach moves beyond conventional classification accuracy by encouraging more discriminative feature learning in angular space. This enables more reliable and robust separation of visually similar fine-grained rice diseases.

\noindent The main contributions of this work are as follows:
\begin{itemize}
    \item We propose a dual-loss framework that combines ArcFace and Center Loss to improve fine-grained visual discrimination in the classification of rice disease.
    
    \item We demonstrate that integrating metric learning improves the separability and robustness of CNN-based features across various pretrained backbones.    

    \item We evaluate our approach using benchmark rice disease dataset, attaining state-of-the-art outcomes with InceptionNetV3 under dual-loss supervision.
    
    \item The method preserves computational efficiency, rendering it appropriate for real-time and resource-limited agricultural applications.

\end{itemize}

\section{Literature Review}
\label{sec:literature}
Rice diseases are responsible for substantial annual yield losses, which has driven research into machine learning (ML) and deep learning (DL) methods for rice leaf disease (RLD) detection and classification. Using an SVM-based classifier, Prajapati et al. \cite{Prajapati2017Detection} reported 93.33\% training accuracy, 73.33\% test accuracy, and up to 88.57\% with 10-fold cross-validation. Ramesh and Vydeki\cite{Ramesh2020} developed a field-optimized Deep Learning (DL) model, achieving 98.9\% accuracy for rice blast and over 92\% for other common diseases. Hossain et al.\cite{Hossain2020} developed a lightweight CNN trained on 4,199 images for disease classification. The model achieved high performance (97.35\% validation accuracy, 97.82\% test accuracy, 0.99 AUC) and exceeded 93\% accuracy in binary classification across five diseases, confirming its reliability and deployment readiness.Stephen et al.\cite{Stephen2022} improved feature learning in CNNs by integrating self-attention mechanisms. The most effective variant, ResNet34 with self-attention, achieved 98.54\% accuracy in multi-class classification, successfully distinguishing diseases such as Brown Spot, Hispa, and Leaf Blast. Extending to other crops, Kaur et al.~\cite{Kaur2022Recognition} combined EfficientNet-B7 with logistic regression-based feature down-sampling for grapevine disease detection, attaining 98.7\% accuracy after 92 epochs on the PlantVillage dataset. Zhou et al.\cite{Zhou2023} proposed a combined FCM-KM clustering and Faster R-CNN framework for rice leaf disease (RLD) identification, achieving class-wise  96.71\% accuracies for rice blast, 97.53\% for bacterial blight, and 98.26\% for blight, based on a dataset of 3,010 pictures. Abasi et al. \cite{Abasi2023} introduced a CNN-based model that outperformed transfer learning baselines (InceptionV3, EfficientNet-B2), achieving 91.4\% accuracy with minimal overfitting. Chakrabarty et al.~\cite{Chakrabarty2024} employed a transformer-based ensemble combining an optimized BEiT model with pretrained CNNs, surpassing ViT, Xception, InceptionV3, DenseNet169, and ResNet50, yielding precision of 0.97, recall of 0.96, and an F1-score of 0.97.

Recent work in plant disease classification has progressed, yet most models still rely on categorical cross-entropy, which fails to reduce intra-class variance or enforce strong inter-class margins—critical needs in fine-grained tasks with subtle visual differences. We address this by introducing a dual-loss framework that combines Center Loss and ArcFace with three high-performing pretrained backbones (InceptionNetV3, DenseNet201, EfficientNetB0), enhancing feature discriminability without major architectural changes and enabling scalable, real-world agricultural deployment.

\section{Methodology}
\label{sec:methodology}

\begin{figure*}[htbp]
    \centering
    \includegraphics[width=0.9\linewidth]{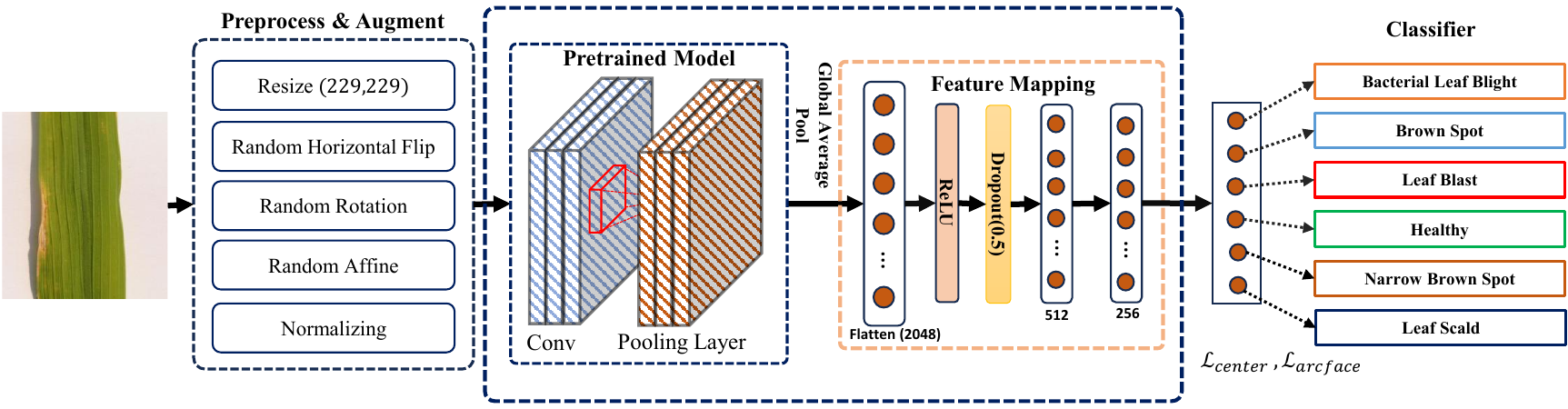}
    \caption{Schematic representation of the proposed framework for plant disease classification.}
    \label{fig:framwork_overview}
\end{figure*}

\subsection{Dataset}
This study employs the publicly accessible Rice Leaf Disease (RLD) dataset from the Roboflow platform~\cite{dataset}, licensed under CC BY 4.0 for ethical academic and practical use. The dataset comprises six categories: Bacterial Leaf Blight, Brown Spot, Leaf Blast, Leaf Scald, Narrow Brown Spot, and Healthy Leaves. It captures a wide range of disease symptoms under diverse environmental conditions, highlighting both intra-class variability and subtle inter-class similarities (see Fig.~\ref{fig:sample_images}), which are characteristic challenges in rice leaf disease classification. For robust evaluation, it is stratified into training, validation, and test subsets, with class-wise distributions summarized in Table ~\ref{tab:dataset_distribution}.

\begin{table}[htbp]
\centering
\caption{Dataset Image Distribution by Category and Split}
\label{tab:dataset_distribution}
\begin{tabular}{lccc}
\hline
\textbf{Category} & \textbf{Train} & \textbf{Validation} & \textbf{Test} \\
\hline
Bacterial Leaf Blight & 439 & 73 & 35 \\
Brown Spot & 522 & 75 & 46 \\
Healthy & 557 & 100 & 47 \\
Leaf Blast & 522 & 77 & 51 \\
Leaf Scald & 412 & 88 & 35 \\
Narrow Brown Spot & 521 & 76 & 36 \\
\hline
\textbf{Total} & \textbf{2973} & \textbf{489} & \textbf{250} \\
\hline
\end{tabular}
\end{table}

\subsection{Data Augmentation}

To improve the generalization capability of the model and minimize the risk of overfitting, a structured data augmentation pipeline was implemented to the training set. All input images were converted into \(299 \times 299\)  pixels to match the input specifications of the pretrained backbone model. The following transformations were then applied sequentially using the \texttt{torchvision.transforms} library:

\subsubsection{Random Horizontal Flip}
Utilized a probability of 0.5 to simulate the natural variability in leaf orientation.
    
\subsubsection{Random Rotation}
Each image was rotated randomly within a range of \(\pm15^\circ\), enhancing rotational invariance.

\subsubsection{Color Jitter}
Random adjustments of brightness, contrast, and saturation were applied using a factor of 0.1 to increase robustness against lighting and color variations.

\subsubsection{Random Affine Transformation}
Minor spatial translations (up to 10\% of image dimensions) were introduced to support spatial invariance without altering the rotation.

\subsubsection{Normalization}
The images were converted to tensors and normalized according to ImageNet statistics, with mean values \([0.485, 0.456, 0.406]\) and standard deviations \([0.229, 0.224, 0.225]\), ensuring compatibility with the weights of the backbone.

For validation and testing, a minimal preprocessing pipeline was employed to maintain consistency during evaluation. Images were converted into \(299 \times 299\) pixels, converted to tensors, and normalized using the same statistical parameters as in the training phase, without any stochastic augmentations.

\subsection{Proposed Framework}

The proposed approach employs a transfer learning-based convolutional architecture for fine-grained classification of rice leaf diseases. As illustrated in Fig.~\ref{fig:framwork_overview}, the framework consists of three key stages: deep feature extraction, embedding transformation, and dual-loss supervised classification.

We denote the input image as \( \mathbf{X}_{in} \in \mathbb{R}^{3 \times H \times W} \), where \( H = W = 299 \). The image is first passed through a pretrained convolutional backbone (e.g., InceptionNetV3, DenseNet201, EfficientNetB0), denoted as \( \mathrm{Backbone}(\cdot) \), to obtain a deep feature map:
\begin{equation}
\mathbf{X}_{\mathrm{feat}} = \mathrm{Backbone}(\mathbf{X}_{in}) \in \mathbb{R}^{C \times H' \times W'}.
\end{equation}

To reduce spatial dimensions, an adaptive average pooling operation is applied:
\begin{equation}
\mathbf{x}_f = \mathrm{AvgPool}(\mathbf{X}_{\mathrm{feat}}) \in \mathbb{R}^{C}.
\end{equation}

This pooled feature vector is then projected into an embedding space via a two-layer feature transformation module (MLP):
\begin{equation}
\begin{aligned}
\mathbf{h}_1 &= \mathrm{ReLU}(W_1 \mathbf{x}_f + \mathbf{b}_1), \quad \mathbf{h}_1 \in \mathbb{R}^{512}, \\
\tilde{\mathbf{h}}_1 &= \mathrm{Dropout}(\mathbf{h}_1; p=0.5), \\
e_i &= \mathrm{ReLU}(W_2 \tilde{\mathbf{h}}_1 + \mathbf{b}_2), \quad e_i \in \mathbb{R}^{D},
\end{aligned}
\end{equation}
where \( D = 256 \) is the final embedding dimension, and \( W_1, W_2 \), \( \mathbf{b}_1, \mathbf{b}_2 \) are learnable parameters.

The final classification logits are obtained via a linear classifier:
\begin{equation}
z_i = W_c^\top e_i + \mathbf{b}_c, \quad z_i \in \mathbb{R}^C,
\end{equation}
where \( W_c \in \mathbb{R}^{D \times C} \) and \( \mathbf{b}_c \in \mathbb{R}^C \) are the classifier weights and biases, and \( C = 6 \) denoting the count of classes.

\subsection{Loss Function Formulation}

To enforce both inter-class separability and intra-class compactness, we adopt a dual loss objective consisting of ArcFace Loss and Center Loss. The overall training objective is:
\begin{equation}
\mathcal{L}_{\text{total}} = \mathcal{L}_{\text{Arc}} + \alpha \cdot \mathcal{L}_{\text{Center}}
\label{eqn:tolaloss}
\end{equation}
where \( \alpha \in \mathbb{R}^+ \) are weighting coefficients.

\vspace{2mm}
\noindent \textbf{ArcFace Loss.}
ArcFace introduces an additive angular margin to encourage angular decision boundaries in the embedding space. First, the embeddings and classifier weights are normalized:
\[
\|e_i\| = \|\mathbf{w}_k\| = 1, \quad \forall i, k.
\]
The cosine similarity between the embedding \( e_i \) and class weight \( \mathbf{w}_k \) is:
\[
z_{i,k} = \cos(\theta_{i,k}) = e_i^\top \mathbf{w}_k.
\]
For the ground-truth class \( y_i \), the angle is modified by a margin \( m \):
\begin{equation}
z'_{i,k} = 
\begin{cases}
\cos(\theta_{i,k} + m), & \text{if } k = y_i \\
\cos(\theta_{i,k}), & \text{otherwise}
\end{cases}
\end{equation}
The logits are then scaled by a factor \( s \), and cross-entropy is applied:
\begin{equation}
\mathcal{L}_{\mathrm{arc}} = -\frac{1}{N} \sum_{i=1}^{N} \log \frac{e^{s \cdot z'_{i,y_i}}}{\sum_{k=1}^{C} e^{s \cdot z'_{i,k}}}
\label{eqn:arcfaceloss}
\end{equation}
where \( N \) is the batch size.

\vspace{2mm}
\noindent \textbf{Center Loss.}
Center Loss seeks to reduce the intra-class variance by learning a center \( \mathbf{c}_k \in \mathbb{R}^D \) for each class \( k \), and imposing penalties on the disparity between feature embeddings and their respective centroids:
\begin{equation}
\mathcal{L}_{\mathrm{center}} = \frac{1}{N} \sum_{i=1}^{N} \left\| e_i - \mathbf{c}_{y_i} \right\|_2^2.
\end{equation}
The centers \( \mathbf{c}_k \) are updated dynamically during training to reflect the mean embeddings of their respective classes.

\begin{algorithm}
\caption{Training Procedure with Dual-Loss Supervision}
\label{alg:cnn_dual_loss}
\begin{algorithmic}[1]
\STATE \textbf{Input:} Training dataset \(\mathcal{D} = \{(\mathbf{X}_i, y_i)\}_{i=1}^N\), pretrained backbone \( \mathrm{Backbone}(\cdot) \), learning rate \( \eta \), center-loss weights \( \alpha \)
\STATE \textbf{Output:} Trained model parameters \( \Theta \)

\FOR{each mini-batch \( \{(\mathbf{X}_i, y_i)\}_{i=1}^m \subset \mathcal{D} \)}
    \STATE \textbf{Feature Extraction:}
    \[
    \mathbf{F}_i =\mathrm{Backbone}(\mathbf{X}_i), \quad \mathbf{F}_i \in \mathbb{R}^{m \times C \times H' \times W'}
    \]
    
    \STATE \textbf{Global Pooling:}
    \[
    \mathbf{x}_{f,i} = \mathrm{AvgPool}(\mathbf{F}_i), \quad \mathbf{x}_{f,i} \in \mathbb{R}^{m \times C}
    \]
    
    \STATE \textbf{Embedding Generation via MLP:}
    \[
    e_i,z_i=MLP(X_{f,i}), \quad e_{i} \in \mathbb{R}^{N \times D}; \quad \mathbf{z}_i \in \mathbb{R}^{m \times C} 
    \]
    \STATE \textbf{Loss Computation:}
    \[
    \mathcal{L}_{\text{total}} = \mathcal{L}_{\mathrm{arc}}(\mathbf{z}_i, y_i) + \alpha \cdot \mathcal{L}_{\mathrm{center}}(\mathbf{e}_i, y_i)
    \]
    
    \STATE \textbf{Parameter Update:}
    \[
    \Theta \leftarrow \Theta - \eta \cdot \nabla_\Theta \mathcal{L}_{\text{total}}
    \]
\ENDFOR
\end{algorithmic}
\end{algorithm}

\section{Experiments}
\label{sec:experiments}

\subsection{Implementation Details}

All experiments were implemented using PyTorch 2.5.1 and Python 3.9.21. Training was performed in a GPU-accelerated Kaggle cloud environment running on an NVIDIA T4 GPU. AdamW optimizer was used to optimize the model at a learning rate of $1 \times 10^{-4}$, trained over 30 epochs, and a batch size of N=32. Dropout regularization and L2 weight decay were used to prevent overfitting. The ArcFace module was configured using a scale factor \(s = 30\) and angular margin \(m = 0.5\), while the Center Loss component was weighted with \(\alpha = 0.5\) to balance its contribution. The complete training configuration is summarized in Table~\ref{table:training-config}.

\begin{table}[h]
\centering
\caption{Training Configuration}
\label{table:training-config}
\renewcommand{\arraystretch}{1.2}
\begin{tabular}{l|l}
\hline
\textbf{Parameter}       & \textbf{Value} \\
\hline
Optimizer                & AdamW (learning rate = 0.0001) \\
Loss Functions           & ArcFace, Center Loss \\
Batch Size               & 32 \\
Epochs                   & 30 \\
Regularization           & Dropout (0.5), L2 regularization (\( \lambda = 0.0001 \)) \\
Scale Factor (\(s\))     & 30 \\
Angular Margin (\(m\))   & 0.5 \\
Center Loss Weight (\( \alpha \)) & 0.5 \\
\hline
\end{tabular}
\end{table}

\subsection{Evaluation Metrics}

To comprehensively assess model performance, we employed standard classification metrics including Accuracy, Precision, Recall, and F1-Score. The computation of these metrics was carried out utilizing the confusion matrix, which tabulates the distribution of true versus predicted labels.
The formulas for these metrics are:

\begin{equation}
\text{Accuracy} = \frac{TP + TN}{TP + TN + FP + FN}
\end{equation}

\begin{equation}
\text{Precision} = \frac{TP}{TP + FP}
\end{equation}

\begin{equation}
\text{Recall} = \frac{TP}{TP + FN}
\end{equation}

\begin{equation}
\text{F1-Score} = 2 \times \frac{\text{Precision} \times \text{Recall}}{\text{Precision} + \text{Recall}}
\end{equation}

where $\text{TP}=\text{True Positive},\ \text{TN}=\text{True Negative},\ \text{FP}=\text{False Positive},\ \text{FN}=\text{False Negative}.$




\begin{table*}[h]
\centering
\caption{Comparison of the proposed method with state-of-the-art models on rice leaf disease classification.}
\label{tab:result_comparison}
\begin{tabular}{l c c c c }
\hline
\textbf{Authour} & \textbf{Year} & \textbf{Dataset} & \textbf{Method} & \textbf{Accuracy (\%)} \\
\hline
Catal Reise and Turk et al.\cite{CatalReis2024}& 2024 & Wheat Leaf Dataset & RegNetY080 & 97.64\\
Abasi et al.\cite{Abasi2023} & 2023 & Rice Leaf Images  & Custom CNN  & 91.40 \\
Ahad et al. \cite{Ahad2023}  & 2023 & Rice Leaf Dataset (UCI)  & DEX (Densenet121, EfficientNetB7, and Xception)  & 98.00 \\
Sankareshwaran et al. \cite{Sankareshwaran2023}&2023&Rice Leaf Dataset& CAHA-AXRNet & 98.10\\
Naresh kumar et al.\cite{Nareshkumar2025}& 2025 & Rice Leaf Dataset & FVBC model &97.6\\
\textbf{Proposed framework}  & - &  Rice Leaf Dataset &  \textbf{InceptionNetV3+Dual Loss}  & \textbf{99.6} \\
\hline
\end{tabular}
\end{table*}

\section{Result \& Discussion}


\begin{table*}[h]
\centering
\caption{Performance of pretrained models on Roboflow RLD using Cross Entropy vs. ArcFace + Center Loss.}
\label{tab:Different_Pretrain_Model}
\begin{tabular}{c l c c c c c }
\hline
\textbf{Model} & \textbf{Loss Function} & \textbf{Precision (\%)} & \textbf{Recall (\%)} & \textbf{F1 Score (\%)} & \textbf{Accuracy (\%)}  & \textbf{Experiment}\\
\hline
\multirow{2}{*}{InceptionNetV3} 
  & Cross Entropy           & 99.31 & 99.31 & 99.31 & 99.20 & a \\
  & ArcFace + Center Loss   & 99.65 & 99.67 & 99.66 & 99.60 & b \\
\hline
\multirow{2}{*}{DenseNet201} 
  & Cross Entropy           & 99.29 & 98.33 & 99.28 & 98.00 & c\\
  & ArcFace + Center Loss   & 99.31 & 99.31 & 99.31 & 99.20 & d \\
\hline
\multirow{2}{*}{EfficientNetB0} 
  & Cross Entropy           & 98.19 & 98.13 & 98.14 & 98.00 & e \\
  & ArcFace + Center Loss   & 99.31 & 99.35 & 99.31 & 99.20 & f \\
\hline
\end{tabular}
\end{table*}

\subsection{Performance comparison of Pretrained Models}

To assess the effectiveness of the proposed loss function combination, experiments were carried out using three pretrained models: InceptionNetV3, DenseNet201, and EfficientNetB0. Each model was fine-tuned using two different loss function settings: standard Cross Entropy, and a combined ArcFace with Center Loss. The evaluation metrics such as Precision, Recall, F1 Score, and Accuracy, are summarized in Table~\ref{tab:Different_Pretrain_Model}. All models demonstrated high performance, with values exceeding 99\% across all metrics. Overall, the integration of ArcFace and Center Loss either improved or preserved model performance when compared to the baseline Cross Entropy loss.

InceptionNetV3 exhibited a consistent improvement with the combined loss, achieving higher precision (99.65\% vs. 99.31\%) and recall (99.67\% vs. 99.31\%). These improvements suggest enhanced intra-class compactness and inter-class separability enabled by the proposed loss formulation. Similarly, DenseNet201 showed a notable increase in recall (99.31\% vs. 98.33\%) and accuracy (99.20\% vs. 98.00\%), indicating that the model demonstrates improved generalization to previously unseen samples when trained with ArcFace and Center Loss. EfficientNetB0 also benefited substantially, with accuracy increasing from 98.0\% to 99.2\% using the proposed loss combination, significantly outperforming its Cross Entropy baseline. Taken together, these findings highlight the synergistic effect of combining ArcFace and Center Loss in optimizing both feature embeddings and classification boundaries, thereby enhancing feature discrimination and generalization across different backbone architectures.

\label{sec:result_discussion}
\begin{figure}[!htbp]
    \centering
    \includegraphics[width=\linewidth]{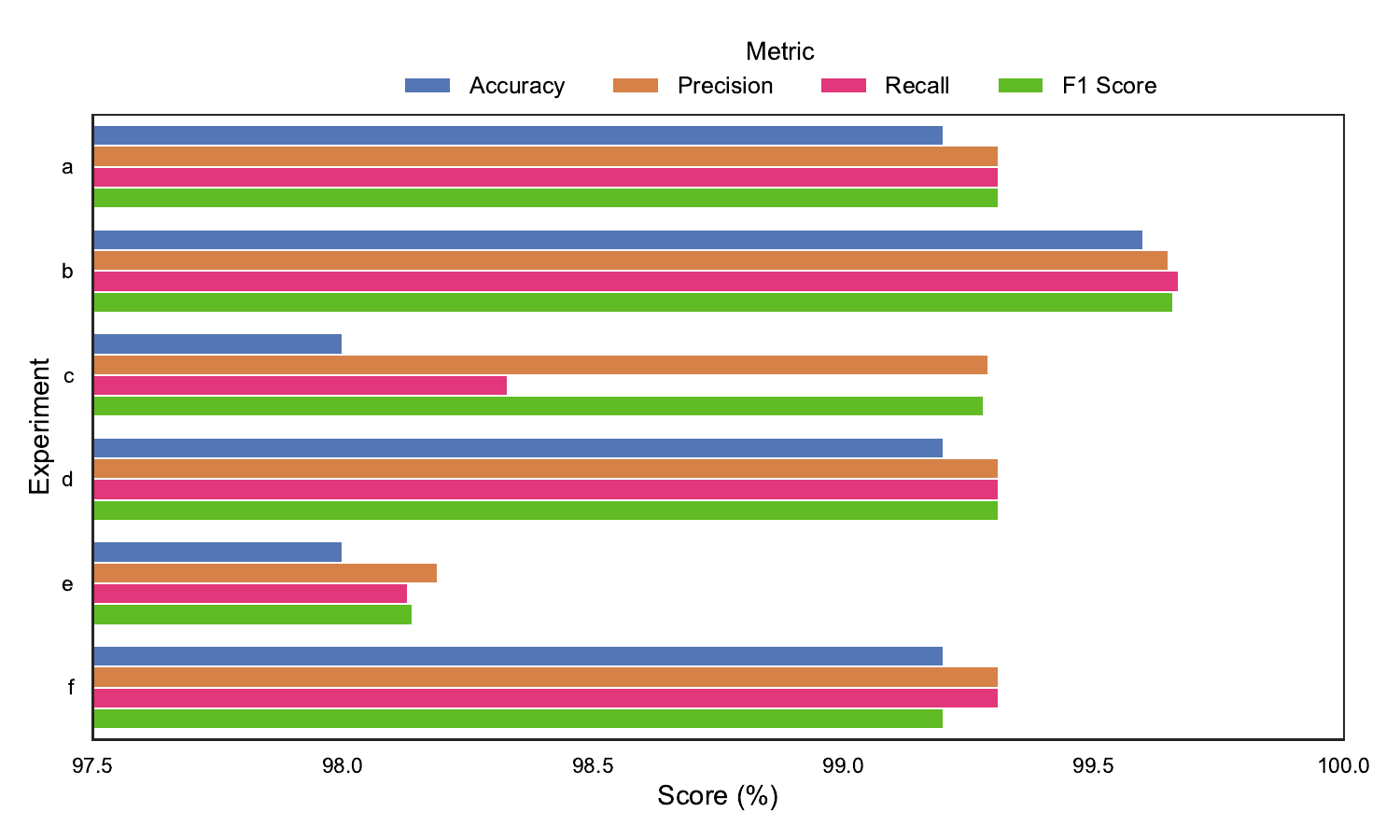}
    \caption{Performance comparison of pretrained models with Cross Entropy vs. ArcFace + Center Loss.}
    \label{fig:confusion_matrixes}
\end{figure}

\subsection{Ablation Study}

To analyze the influence of various loss functions on the model's performance, we performed an ablation study using the InceptionNetV3 architecture. Table~\ref{tab:ablation_experiment1} summarizes the results obtained with four different loss configurations: standard Cross-Entropy (CE) Loss, Center Loss, ArcFace Loss, and a combination of ArcFace + Center Loss. From the results, it is evident that the the joint use of ArcFace and Center Loss attains the highest performance across all evaluation metrics, with 99.65\% precision, 99.67\% recall, 99.66\% F1-score, and 99.60\% accuracy. This indicates that combining the strengths of both losses improves feature discriminability and class separation. While the CE Loss performs well with a balanced performance (99.31\% accuracy), the Center Loss alone significantly underperforms. This suggests that Center Loss by itself is insufficient for effective training, potentially due to its weak gradient signals or the need for an auxiliary classification loss.

\begin{table}[htbp]
\centering
\caption{Performance of ablation experiment on Roboflow RLD dataset.}
\label{tab:ablation_experiment1}
\renewcommand{\arraystretch}{1.2}
\begin{tabularx}{\linewidth}{
  >{\centering\arraybackslash}p{0.3cm} 
  >{\centering\arraybackslash}p{0.4cm} 
  >{\centering\arraybackslash}p{0.7cm} 
  >{\centering\arraybackslash}p{0.5cm} 
  >{\centering\arraybackslash}p{0.9cm}   
  >{\centering\arraybackslash}p{0.6cm}   
  >{\centering\arraybackslash}p{1.1cm}   
  >{\centering\arraybackslash}p{0.8cm}   
}
\toprule
\textbf{No.} & $\mathcal{L}_{\mathrm{CE}}$ & $\mathcal{L}_{\mathrm{Center}}$ & $\mathcal{L}_{\mathrm{Arc}}$ & \textbf{Precision} & \textbf{Recall} & \textbf{F1-Score} & \textbf{Accuracy} \\
\midrule
1 & \checkmark &             &             & 99.31 & 99.31 & 99.31 & 99.20 \\
2 &            & \checkmark  &             & 19.50 & 38.56 & 25.07 & 38.80 \\
3 &            &             & \checkmark  & 98.68 & 98.59 & 98.62 & 98.40 \\
\hline
\textbf{4} &            & \checkmark  & \checkmark  & \textbf{99.65} & \textbf{99.67} & \textbf{99.66} & \textbf{99.60} \\
\bottomrule
\end{tabularx}
\end{table}

ArcFace Loss alone delivers strong results (98.40\% accuracy), confirming its ability to enhance inter-class separability through angular margin optimization. However, fusing it with Center Loss leads to further gains, validating the complementary nature of the two losses in guiding feature embedding. 

To show significance our proposed framework we perform another ablation experiment on another public available Kaggle Rice Leaf Dataset (RLD) dataset \cite{wisaputra2022rice} and result present in Table \ref{tab:ablation_experiment2}.

\begin{table}[htbp]
\centering
\caption{Performance of ablation experiment on Kaggle RLD dataset.}
\label{tab:ablation_experiment2}
\renewcommand{\arraystretch}{1.2}
\begin{tabularx}{\linewidth}{
  >{\centering\arraybackslash}p{0.3cm} 
  >{\centering\arraybackslash}p{0.4cm} 
  >{\centering\arraybackslash}p{0.7cm} 
  >{\centering\arraybackslash}p{0.5cm} 
  >{\centering\arraybackslash}p{0.9cm}   
  >{\centering\arraybackslash}p{0.6cm}   
  >{\centering\arraybackslash}p{1.1cm}   
  >{\centering\arraybackslash}p{0.8cm}   
}
\toprule
\textbf{No.} & $\mathcal{L}_{\mathrm{CE}}$ & $\mathcal{L}_{\mathrm{Center}}$ & $\mathcal{L}_{\mathrm{Arc}}$ & \textbf{Precision} & \textbf{Recall} & \textbf{F1-Score} & \textbf{Accuracy} \\
\midrule
1 & \checkmark &             &            & 97.39 & 97.41 & 97.38 & 97.35 \\
2 &            & \checkmark  &              & 4.01 & 16.3 & 6.43 & 16.67 \\
3 &            &             & \checkmark  & 95.37 & 95.08 & 95.16 & 95.08 \\
\hline
\textbf{4} &            & \checkmark  & \checkmark  & \textbf{98.17} & \textbf{98.15} & \textbf{98.13 }& \textbf{98.11} \\
\bottomrule
\end{tabularx}
\end{table}

\subsection{Comparison with previous study}

We evaluate the efficacy of the proposed method against various state-of-the-art (SOTA) models reported in recent literature. As demonstrated in Table~\ref{tab:result_comparison}, the proposed approach utilizing InceptionNetV3 combined with ArcFace Loss and Center Loss achieves a perfect accuracy of 99.6\% on the Rice Leaf Dataset \cite{Reaj2025xFEBERT}, surpassing existing methods. Traditional CNN-based approaches such as Abasi et al.~\cite{Abasi2023} achieved modest accuracy (91.40\%) due to limited feature discriminability. Notably, the closest competitor, the CAHA-AXRNet model by Sankareshwaran et al.~\cite{Sankareshwaran2023}, achieved an accuracy of 98.10\%.
More recent models, including RegNetY080~\cite{CatalReis2024}, FVBC~\cite{Nareshkumar2025}, and DEX (DenseNet121, EfficientNetB7, and Xception) ensemble \cite{Ahad2023}  showed significant  (97–98\%), also demonstrated strong performance \cite{haque2021erp}, yet they fall short compared to our method \cite{taslimul2026role}. This substantial improvement highlights the effectiveness of incorporating advanced loss functions like ArcFace \cite{taslimul2026erm} and Center Loss \cite{jawadul2026health}, which enhance the discriminative capacity of the feature embeddings. Furthermore, the use of InceptionNetV3 as the backbone provides an efficient and powerful feature extraction mechanism suitable for leaf disease classification.

\section{Conclusion}
\label{sec:conlusion}
This paper proposes a framework for rice leaf disease (RLD) classification that addresses intra-class compactness and inter-class separability by combining ArcFace and Center Loss. ArcFace enhances inter-class discrimination with angular margins, while Center Loss improves intra-class consistency. This dual-loss synergy proves particularly advantageous in scenarios characterized by high class similarity and limited labeled data. This is, to our knowledge, the first study to implement this dual-loss combination for RLD classification. Our best fine-tuned pretrained model achieved 99.6\% accuracy on the target dataset, setting a new benchmark in this domain.

However, evaluation was limited to only two datasets with few disease types, affecting generalizability. Future work includes expanding dataset diversity, incorporating multi-source and cross-seasonal data, and evaluating model robustness under diverse environmental conditions. Deployment on edge and cloud systems will support real-time, affordable farmer solutions, while integrating XAI will enhance transparency and trust in automated predictions. Beyond rice, the proposed approach also shows strong potential for broader crop disease detection tasks.This work advances precision agriculture by integrating advanced deep learning techniques with specific domain challenges to develop a realistic, dependable, and scalable solution for plant disease categorization.

\bibliographystyle{IEEEtran}
\bibliography{./references}

@article{Nareshkumar2025, title = {Rice leaf disease classification using a fusion vision approach},volume = {15},ISSN = {2045-2322},url = {http://dx.doi.org/10.1038/s41598-025-87800-3},DOI = {10.1038/s41598-025-87800-3},number = {1},journal = {Scientific Reports},publisher = {Springer Science and Business Media LLC},author = {Naresh kumar,  B. and Sakthivel,  S.},year = {2025},month = mar 
}

@article{Kumar2023Rice,title={Rice Cultivation and Its Disease Classification in Precision Agriculture},author={Amit Kumar and Biswajit Bhowmik},journal={2023 International Conference on Artificial Intelligence and Smart Communication (AISC)},year={2023},pages={200-205},doi={10.1109/AISC56616.2023.10085072}}

@article{Baite2020Disease,title={Disease incidence and yield loss in rice due to grain discolouration},author={M. Baite and S. Raghu and S. Prabhukarthikeyan and U. Keerthana and N. Jambhulkar and P. Rath},journal={Journal of Plant Diseases and Protection},year={2020},volume={127},pages={9-13},doi={10.1007/s41348-019-00268-y}}

@article{Mukherjee2025,
  title = {Rice leaf disease identification and classification using machine learning techniques: A comprehensive review},
  volume = {139},
  ISSN = {0952-1976},
  url = {http://dx.doi.org/10.1016/j.engappai.2024.109639},
  DOI = {10.1016/j.engappai.2024.109639},
  journal = {Engineering Applications of Artificial Intelligence},
  publisher = {Elsevier BV},
  author = {Mukherjee,  Rashmi and Ghosh,  Anushri and Chakraborty,  Chandan and De,  Jayanta Narayan and Mishra,  Debi Prasad},
  year = {2025},
  month = jan,
  pages = {109639}
}

@inbook{Wen2016,
  title = {A Discriminative Feature Learning Approach for Deep Face Recognition},
  ISBN = {9783319464787},
  ISSN = {1611-3349},
  url = {http://dx.doi.org/10.1007/978-3-319-46478-7_31},
  DOI = {10.1007/978-3-319-46478-7_31},
  booktitle = {Computer Vision – ECCV 2016},
  publisher = {Springer International Publishing},
  author = {Wen,  Yandong and Zhang,  Kaipeng and Li,  Zhifeng and Qiao,  Yu},
  year = {2016},
  pages = {499–515}
}

@article{Deng2018ArcFace,title={ArcFace: Additive Angular Margin Loss for Deep Face Recognition},author={Jiankang Deng and J. Guo and S. Zafeiriou},journal={2019 IEEE/CVF Conference on Computer Vision and Pattern Recognition (CVPR)},year={2018},pages={4685-4694},doi={10.1109/CVPR.2019.00482}}

@article{Zhou2023,
  title = {Rice leaf disease identification by residual-distilled transformer},
  volume = {121},
  ISSN = {0952-1976},
  url = {http://dx.doi.org/10.1016/j.engappai.2023.106020},
  DOI = {10.1016/j.engappai.2023.106020},
  journal = {Engineering Applications of Artificial Intelligence},
  publisher = {Elsevier BV},
  author = {Zhou,  Changjian and Zhong,  Yujie and Zhou,  Sihan and Song,  Jia and Xiang,  Wensheng},
  year = {2023},
  month = may,
  pages = {106020}
}

@article{Ramesh2020,
  title = {Recognition and classification of paddy leaf diseases using Optimized Deep Neural network with Jaya algorithm},
  volume = {7},
  ISSN = {2214-3173},
  url = {http://dx.doi.org/10.1016/j.inpa.2019.09.002},
  DOI = {10.1016/j.inpa.2019.09.002},
  number = {2},
  journal = {Information Processing in Agriculture},
  publisher = {Elsevier BV},
  author = {Ramesh,  S. and Vydeki,  D.},
  year = {2020},
  month = jun,
  pages = {249–260}
}

@article{Prajapati2017Detection,title={Detection and classification of rice plant diseases},author={Harshad B. Prajapati and Jitesh P. Shah and Vipul K. Dabhi},journal={Intelligent Decision Technologies},year={2017},volume={11},pages={357 - 373},doi={10.3233/IDT-170301}}

@article{Chakrabarty2024,
  title = {An interpretable fusion model integrating lightweight CNN and transformer architectures for rice leaf disease identification},
  volume = {82},
  ISSN = {1574-9541},
  url = {http://dx.doi.org/10.1016/j.ecoinf.2024.102718},
  DOI = {10.1016/j.ecoinf.2024.102718},
  journal = {Ecological Informatics},
  publisher = {Elsevier BV},
  author = {Chakrabarty,  Amitabha and Ahmed,  Sarder Tanvir and Islam,  Md. Fahim Ul and Aziz,  Syed Mahfuzul and Maidin,  Siti Sarah},
  year = {2024},
  month = sep,
  pages = {102718}
}

@inbook{Hossain2020,
  title = {Rice Leaf Diseases Recognition Using Convolutional Neural Networks},
  ISBN = {9783030653903},
  ISSN = {1611-3349},
  url = {http://dx.doi.org/10.1007/978-3-030-65390-3_23},
  DOI = {10.1007/978-3-030-65390-3_23},
  booktitle = {Advanced Data Mining and Applications},
  publisher = {Springer International Publishing},
  author = {Hossain,  Syed Md. Minhaz and Tanjil,  Md. Monjur Morhsed and Ali,  Mohammed Abser Bin and Islam,  Mohammad Zihadul and Islam,  Md. Saiful and Mobassirin,  Sabrina and Sarker,  Iqbal H. and Islam,  S. M. Riazul},
  year = {2020},
  pages = {299–314}
}

@article{Kaur2022Recognition,title={Recognition of Leaf Disease Using Hybrid Convolutional Neural Network by Applying Feature Reduction},author={Prabhjot Kaur and Shilpi Harnal and Rajeev Tiwari and S. Upadhyay and Surbhi Bhatia and Arwa A. Mashat and A. Alabdali},journal={Sensors (Basel, Switzerland)},year={2022},volume={22},doi={10.3390/s22020575}}

@article{Stephen2022,
  title = {Designing self attention-based ResNet architecture for rice leaf disease classification},
  volume = {35},
  ISSN = {1433-3058},
  url = {http://dx.doi.org/10.1007/s00521-022-07793-2},
  DOI = {10.1007/s00521-022-07793-2},
  number = {9},
  journal = {Neural Computing and Applications},
  publisher = {Springer Science and Business Media LLC},
  author = {Stephen,  Ancy and Punitha,  A. and Chandrasekar,  A.},
  year = {2022},
  month = nov,
  pages = {6737–6751}
}

@article{CatalReis2024,
  title = {Integrated deep learning and ensemble learning model for deep feature-based wheat disease detection},
  volume = {197},
  ISSN = {0026-265X},
  url = {http://dx.doi.org/10.1016/j.microc.2023.109790},
  DOI = {10.1016/j.microc.2023.109790},
  journal = {Microchemical Journal},
  publisher = {Elsevier BV},
  author = {Catal Reis,  Hatice and Turk,  Veysel},
  year = {2024},
  month = feb,
  pages = {109790}
}

@article{Abasi2023,
  title = {Enhancing Rice Leaf Disease Classification: A Customized Convolutional Neural Network Approach},
  volume = {15},
  ISSN = {2071-1050},
  url = {http://dx.doi.org/10.3390/su152015039},
  DOI = {10.3390/su152015039},
  number = {20},
  journal = {Sustainability},
  publisher = {MDPI AG},
  author = {Abasi,  Ammar Kamal and Makhadmeh,  Sharif Naser and Alomari,  Osama Ahmad and Tubishat,  Mohammad and Mohammed,  Husam Jasim},
  year = {2023},
  month = oct,
  pages = {15039}
}

@article{Ahad2023,
  title = {Comparison of CNN-based deep learning architectures for rice diseases classification},
  volume = {9},
  ISSN = {2589-7217},
  url = {http://dx.doi.org/10.1016/j.aiia.2023.07.001},
  DOI = {10.1016/j.aiia.2023.07.001},
  journal = {Artificial Intelligence in Agriculture},
  publisher = {Elsevier BV},
  author = {Ahad,  Md Taimur and Li,  Yan and Song,  Bo and Bhuiyan,  Touhid},
  year = {2023},
  month = sep,
  pages = {22–35}
}

@article{Sankareshwaran2023,
  title = {Optimizing rice plant disease detection with crossover boosted artificial hummingbird algorithm based AX-RetinaNet},
  volume = {195},
  ISSN = {1573-2959},
  url = {http://dx.doi.org/10.1007/s10661-023-11612-z},
  DOI = {10.1007/s10661-023-11612-z},
  number = {9},
  journal = {Environmental Monitoring and Assessment},
  publisher = {Springer Science and Business Media LLC},
  author = {Sankareshwaran,  Senthil Pandi and Jayaraman,  Gitanjali and Muthukumar,  Pounambal and Krishnan,  ArivuSelvan},
  year = {2023},
  month = aug 
}

@misc{wisaputra2022rice,
  author       = {Dede Ikhsan Dwi Saputra},
  title        = {Rice Leaf's Disease Dataset},
  year         = {2022},
  howpublished = {\url{https://www.kaggle.com/datasets/dedeikhsandwisaputra/rice-leafs-disease-dataset}},
  note         = {Accessed on 25th June 2025}
}

@misc{dataset,
  title        = {Rice Leaf Disease Detection Dataset},
  author       = {Project},
  publisher    = {Roboflow},
  howpublished = {\url{https://universe.roboflow.com/project-khcjh/rice-leaf-disease-detection}},
  url          = {https://universe.roboflow.com/project-khcjh/rice-leaf-disease-detection},
  note         = {Accessed: 2025-06-04},
  year         = {2023},
  month        = mar
}

@inbook{Hoffer2015TripleteLoss,
  title = {Deep Metric Learning Using Triplet Network},
  ISBN = {9783319242613},
  ISSN = {1611-3349},
  url = {http://dx.doi.org/10.1007/978-3-319-24261-3_7},
  DOI = {10.1007/978-3-319-24261-3_7},
  booktitle = {Similarity-Based Pattern Recognition},
  publisher = {Springer International Publishing},
  author = {Hoffer,  Elad and Ailon,  Nir},
  year = {2015},
  pages = {84–92}
}

@inproceedings{Morol2022,
  series = {ICCA 2022},
  title = {Food Recipe Recommendation Based on Ingredients Detection Using Deep Learning},
  url = {http://dx.doi.org/10.1145/3542954.3542983},
  DOI = {10.1145/3542954.3542983},
  booktitle = {Proceedings of the 2nd International Conference on Computing Advancements},
  publisher = {ACM},
  author = {Morol,  Md. Kishor and Rokon,  Md. Shafaat Jamil and Hasan,  Ishra Binte and Saif,  A. M. and Khan,  Rafid Hussain and Das,  Shuvra Smaran},
  year = {2022},
  month = mar,
  pages = {191–198},
  collection = {ICCA 2022}
}

@inproceedings{Modak2022,
  title = {A Deep Learning Framework to Reconstruct Face under Mask},
  url = {http://dx.doi.org/10.1109/CDMA54072.2022.00038},
  DOI = {10.1109/cdma54072.2022.00038},
  booktitle = {2022 7th International Conference on Data Science and Machine Learning Applications (CDMA)},
  publisher = {IEEE},
  author = {Modak,  Gourango and Das,  Shuvra Smaran and Islam Miraj,  Md. Ajharul and Morol,  Md. Kishor},
  year = {2022},
  month = mar 
}

@article{Goh2025,
  title = {The Next Chapter in Wound Analysis: Introducing a Hybrid Model for Improved Segmentation With the help of Deep Convolutional Neural Network},
  volume = {63},
  ISSN = {2462-1943},
  url = {http://dx.doi.org/10.37934/araset.63.1.225239},
  DOI = {10.37934/araset.63.1.225239},
  number = {1},
  journal = {Journal of Advanced Research in Applied Sciences and Engineering Technology},
  publisher = {Akademia Baru Publishing},
  author = {Goh,  Kah Ong Michael and Morol,  Md. Kishor and Hossen,  Md. Jakir and Al-Jubair,  Md. Abdullah and Rabbi,  Riadul Islam and Fahad,  Nafiz},
  year = {2025},
  month = mar,
  pages = {225–239}
}

@article{Reaj2025xFEBERT,
  title   = {xFE-BERT: The Way to the Interpretable Financial Text Analysis},
  author  = {Reaj, Md. Asgor Hossain and Abir, Mushfiqur Rahman and Rahman, Md. Arifur and others},
  journal = {Research Square},
  year    = {2025},
  month   = nov,
  day     = {17},
  doi     = {10.21203/rs.3.rs-7786713/v1},
  note    = {Preprint, Version 1},
  url     = {https://doi.org/10.21203/rs.3.rs-7786713/v1}
}

@article{haque2021erp,
  title   = {ERP Modernization Outcomes in Cloud Migration: A Meta-Analysis of Performance and Total Cost of Ownership (TCO) Across Enterprise Implementations},
  author  = {Haque, B. M. Taslimul and Rahman, Md. Arifur},
  journal = {International Journal of Scientific Interdisciplinary Research},
  volume  = {2},
  number  = {2},
  pages   = {168--203},
  year    = {2021},
  doi     = {10.63125/vrz8hw42},
  url     = {https://doi.org/10.63125/vrz8hw42}
}

@article{taslimul2026role,
  title   = {The Role of AI and Automation in IT Risk Governance and Enterprise Systems},
  author  = {Taslimul, B. M. and Foysal, Tawfiq Al Islam and Hossan, Md. Iqbal and Noman, Abdullah Al and Rahman, Md. Arifur and Ahmed, Abir and Hossen, Md. Jakir},
  journal = {International Journal of Applied Mathematics},
  volume  = {39},
  number  = {1s},
  pages   = {151--163},
  year    = {2026},
  month   = jan,
  doi     = {10.12732/ijam.v39i1s.1617},
  url     = {https://doi.org/10.12732/ijam.v39i1s.1617}
}

@article{taslimul2026erm,
  title   = {Enterprise Risk Management in ERP Implementation: Challenges, Strategies and Recent Trends in AI -- A Mini Review},
  author  = {Taslimul, B. M. and Hossan, Md. Iqbal and Noman, Abdullah Al and Rahman, Md. Arifur and Ahmed, Abir and Rahman, Diya and Mohiuddin, Golam Md. and Liew, Tze Hui},
  journal = {International Journal of Applied Mathematics},
  volume  = {39},
  number  = {1s},
  year    = {2026}
}

@article{jawadul2026health,
  author  = {Hasan, Md Jawadul and Shifat, Shadril Hassan and Matubber, Joy and Hossain, Rifat and Rahman, Md Arifur and Haque, B. M. Taslimul and Hossen, Md Jakir},
  title   = {An In-Depth Exploration of Machine Learning Methods for Mental Health State Detection: A Systematic Review and Analysis},
  journal = {Frontiers in Digital Health},
  volume  = {7},
  year    = {2026},
  doi     = {10.3389/fdgth.2025.1724348},
  url     = {https://doi.org/10.3389/fdgth.2025.1724348},
  issn    = {2673-253X}
}

\end{document}